%% file: main.tex
\documentclass{article}

\usepackage[final]{corl_2020} % Uncomment for the camera-ready ``final'' version.
\usepackage{graphicx}
\usepackage{graphics} % for pdf, bitmapped graphics files
\usepackage{mathptmx} % assumes new font selection scheme installed
\usepackage{times} % assumes new font selection scheme installed
\usepackage{amsmath} % assumes amsmath package installed
\usepackage{amssymb}  % assumes amsmath package installed
\usepackage{xcolor}
\usepackage{subcaption} %{subfigure}
\usepackage{wrapfig}
\usepackage{lipsum}

\usepackage{tikz}
\usetikzlibrary{shapes,arrows}
\usepackage{amsmath,bm,times}
\usepackage{amssymb}
\usepackage{verbatim}
\usetikzlibrary{calc,positioning,shapes.geometric}
\usetikzlibrary{backgrounds,fit,bayesnet,arrows}

\usepackage{pgfplotstable}
\usepackage{booktabs} % for better looking table creations, but bad with vertical lines by design (package creator despises vertical lines)
\usepackage{rotating}
\usepackage{multirow} % makes it possible to have bigger cells over multiple rows in a table
\usepackage{array} % different options for table cell orientation

\usepackage{algorithm}
\usepackage{algorithmic}

\newcommand{\learnedCost}{C_\psi}
\newcommand{\irlloss}{\mathcal{L}_{IRL}}

\newcommand{\com}[1]{{\textcolor[rgb]{0.502, 0.502, 0.502}{#1}}}

\newcommand\nsincosdemoone{1}
\newcommand\nsincosdemo{10}

\renewcommand\footnotemark{}

\title{Model-Based Inverse Reinforcement Learning from Visual Demonstrations}

\author{
  Neha Das$^{1,*}$\\
  \texttt{neha.das@tum.de} \\
  \And
  Sarah Bechtle$^{1,2,*}$\\
  \texttt{sbechtle@tuebingen.mpg.de} \\
  \And
  Todor Davchev$^{3}$\\
  \texttt{t.b.davchev@ed.ac.uk} \\
  \AND
  Dinesh Jayaraman$^{1,4}$ \\
  \texttt{dineshj@seas.upenn.edu} \\
  \And
  Akshara Rai$^{1}$ \\
  \texttt{akshararai@fb.com} \\
  \And
  Franziska Meier$^{1}$ \\
  \texttt{fmeier@fb.com} 
  \thanks{$^1$Facebook AI Research, $^2$MPI for Intelligent Systems, $^3$University of Edinburgh, $^4$University of Pennsylvania, $^{*}$work done while at FAIR}
}
\pgfplotsset{compat=1.16}
\begin{document}
\maketitle

%===============================================================================

\begin{abstract}
    Scaling model-based inverse reinforcement learning (IRL) to real robotic manipulation tasks with unknown dynamics remains an open problem. The key challenges lie in learning good dynamics models, developing algorithms that scale to high-dimensional state-spaces and being able to learn from both visual and proprioceptive demonstrations. In this work, we present a gradient-based inverse reinforcement learning framework that utilizes a pre-trained visual dynamics model to learn cost functions when given only visual human demonstrations. The learned cost functions are then used to reproduce the demonstrated behavior via visual model predictive control. We evaluate our framework on hardware on two basic object manipulation tasks.
\end{abstract}


%===============================================================================

\section{Introduction}

Learning from demonstrations is a very active area of research, motivated by enabling robots to bootstrap their learning processes. Demonstrations can help in various ways, for instance via inverse reinforcement learning (IRL), where the robot tries to infer the reward or goals from the human demonstrator. Most IRL approaches require demonstrations that couple action and state measurements, which are often costly to acquire. 

In this work, we take a step towards model-based inverse reinforcement learning from visual demonstrations for simple object manipulation tasks. Model-based IRL approaches are thought to be more sample-efficient and hold promises for easier generalization \cite{osa2018algorithmic}. Yet, thus far, their model-free counter-parts have been more successful in real world robotics applications with unknown dynamics \cite{kalakrishnan_2013_irl, boularias2011relativeirl, finn2016guidedirl}.
Several major challenges remain for model-based IRL:
Model-based inverse reinforcement learning comprises two nested optimization problems, an inner and outer optimization step. The inner optimization problem optimizes a policy given a cost function and transition model. Most prior work \cite{abbeel2004apprenticeship, englert2017inverseRL, wulfmeier2017large} assume this transition model (of the environment and the robot) to be known, however the robot typically does not have access to such a model. The outer optimization step aims to optimize the cost function such that the inner step optimizes a policy that matches well with the observed demonstrations. This step is extremely challenging, as it requires measuring the effect of changes in cost function parameters on resulting policy parameters. Prior work \cite{abbeel2004apprenticeship, abbeel2010autonomous, osa2018algorithmic}, approximate this optimization step by minimizing a hand-designed distance metric between demonstrations and policy rollouts. While this approximation makes the outer optimization step practical, it can lead to instabilities in cost function learning.

%i.e. the robot knows how its actions change the environment state. However, when manipulating objects, the robot typically does not have access to such a model. Furthermore, when optimizing cost function parameters, prior work requires a hand-designed feature space in which the distance between demonstration and predicted rollout is minimized. We, in contrast, aim to optimize cost function parameters such that the distance between demonstration and rollout is minimized in the visual observation space. 

%prior model-based IRL algorithms optimize cost functions parameters such that the distance of rollout and demonstration is miminized in a hand-designed feature space. %Furthermore, IRL approaches typically involve a computationally intensive optimization procedure with an outer loop that estimates new cost function parameters, and an inner loop that solves the RL problem given the new cost function. The results in algorithms that do not scale well. 

Our work makes contributions that address these challenges to enable model-based IRL from visual demonstrations: 1) We train keypoint detectors \cite{minderer2019unsupervised} which extract low-dimensional vision features both on human demonstrations, as well as on the robot and pre-train a dynamics model with which the robot can predict how its actions change this low-dimensional feature representation. Once the robot has observed a latent state trajectory from a human demonstration, it can use its own dynamics model to optimize its actions to achieve the same (relative) latent-state trajectory. 2) We introduce a novel inverse reinforcement learning algorithm that enables learning cost functions by differentiating through the inner optimization step. Specifically, our IRL algorithm builds on recent progress in gradient-based bi-level optimization \cite{higher}, which allows us to compute gradients of cost function parameters as a function of the inner loop policy optimization step, leading to more stable and effective optimization. We evaluate our approach by collecting human demonstrations for two basic object manipulation tasks, learn the cost functions for these tasks and reproduce similar behaviors on a Kuka iiwa.

%===============================================================================
\section{Background and Related Work}
\label{sec:background}
The proposed framework builds upon approaches from visual model-predictive control and IRL. This section provides an overview of the related methods and positions our work in context.
\subsection{Visual Model Predictive Control}

At the core of this paper lies the ability to optimize action sequences that minimize a task cost under a given visual dynamics model. Here we highlight how related work optimizes such action sequences and what cost function representations were chosen.
Most approaches either learn a transition model directly in pixel space or jointly learn a latent-space encoding and a dynamics model in that space. For instance, \cite{finn2017deep,  ebert2018_visual_foresight} learn pixel-level transition models and present methods for designing cost functions that evaluate progress to goal pixel positions, registration to goal images, and success classifiers. They optimize action sequences by utilizing the cross entropy method \cite{CEM}. \cite{byravan_2018_se3net_control} map visual observations to a learned pose space, and learn deep dynamics model that predicts changes in that latent space, which is then used to optimize actions using gradient based methods. \cite{watter2015embed} learns locally linear dynamics models from images, and use stochastic optimal control algorithms in conjunction with quadratic cost functions (that penalize distances in latent space). \cite{minderer2019unsupervised, manuelli2019kpam, Kulkarni2019-wz, DIGIT} learn keypoint representations from visual observations, an estimate a transition model in that learned latent state. In this work, we train 2-D keypoint representations of images via self-supervised training similar to \cite{minderer2019unsupervised, Kulkarni2019-wz, DIGIT}. Next, we train a dynamics model in that latent-space and optimize actions via gradient-based methods, similar to \cite{byravan_2018_se3net_control}. This differentiable action optimization is key to our IRL approach as it allows us to realize a fully differentiable inner optimization step  (Section~\ref{sec:learning_cost_funtions}).

\subsection{Inverse Reinforcement Learning}
Scaling inverse reinforcement learning to manipulation tasks in the physical world has proven difficult.
%is by no means trivial. 
This section provides an overview of some of the previously proposed methods, and positions our work in context. Model-free inverse reinforcement learning algorithms have been shown some success on real robotic platforms for manipulation tasks \cite{kalakrishnan_2013_irl, boularias2011relativeirl, finn2016guidedirl}. \citet{kalakrishnan_2013_irl} and \citet{boularias2011relativeirl} only utilize proprioceptive state measurements and do not consider visual feature spaces. 

However, most model-based IRL methods have been limited to simulation settings with known models \cite{abbeel2004apprenticeship, ziebart2008maximum}, and real robotics tasks with known models \cite{englert2017inverseRL}. An exception is the work of \citet{abbeel2010autonomous} that learns dynamics models for helicopter flight tasks and then learns cost functions via apprenticeship learning \cite{abbeel2004apprenticeship}. Constrained optimization methods are a popular choice for IRL approaches \cite{englert2017inverseRL, zhaoinverse, ratliff2009learning}. Scaling such methods to image-based tasks is highly non-trivial. In contrast, we pre-train a visual dynamics model, and present a gradient-based IRL approach, which is built on recent successes in gradient-based bi-level optimization \cite{higher, bechtle2019meta}. 

\paragraph{IRL and Inverse Optimal Control (IOC) from Visual Demonstrations: }
There have been several approaches that utilize visual demonstrations to learn cost functions \cite{wulfmeier2017large, lee2020approximate, sermanet2016unsupervised, finn2016guidedirl}.
\citep{wulfmeier2017large, lee2020approximate} learn cost functions for path planning tasks in urban and track environments, while \citet{sermanet2016unsupervised} and \citet{finn2016guidedirl} focus on manipulation tasks. \cite{sermanet2016unsupervised, finn2016guidedirl} employ a model-free IRL approach to learn reward functions from visual demonstrations. 
Both methods rely on kinesthetic demonstrations, either for the full IOC approach \cite{finn2016guidedirl}; or to initialize the policy that optimizes the learned reward function \cite{sermanet2016unsupervised}.
In contrast, our approach is model-based. We only utilize expert demonstrations as part of the dynamics model training, and can extract cost functions from visual demonstrations only. When optimizing our policies we do not require expert data for initialization.

%===============================================================================
\section{Gradient-Based Visual Model Predictive Control Framework}
\label{sec:system_overview}
\begin{figure}[H]
    \centering
    \includegraphics[width=0.9\textwidth]{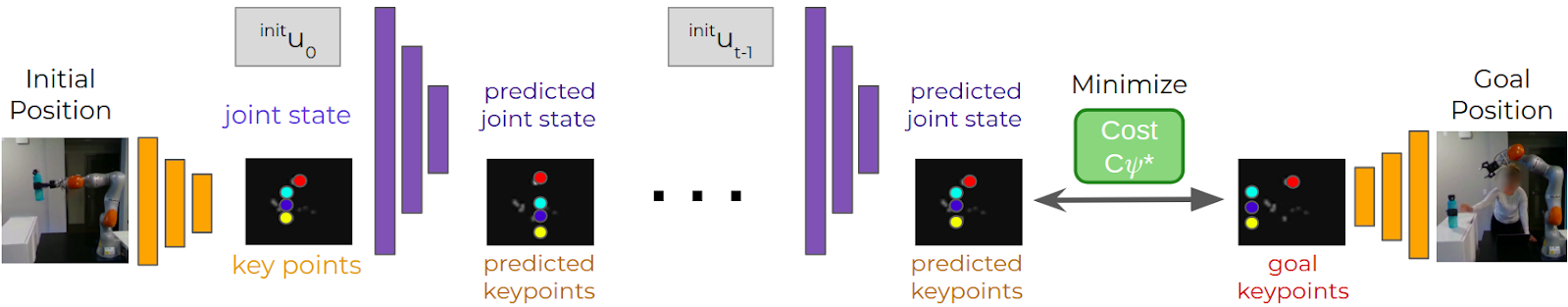}
    \caption{\small Overview of our keypoint-based visual model predictive control framework. Actions are optimized via gradient descent on the cost function.}
    \label{fig:visual_mpc_overview}
    \vspace{-0.3cm}
\end{figure}
In this section we describe our gradient-based visual model predictive control approach that combines recent advances in unsupervised keypoint representations and model-based planning. In the next section, we will build our novel inverse reinforcement learning system on top of this foundation.

The proposed system, depicted in Figure~\ref{fig:visual_mpc_overview}, comprises of following modules: 1) a keypoint detector that produces low-dimensional visual representations, in the form of keypoints, from RGB image inputs; 2) a dynamics model that takes in the current joint state $\theta, \dot{\theta}$ and actions $u$ and predicts the keypoints and joint state at the next time step; and 3) a gradient based visual model-predictive planner that, given the dynamics model and a cost function, optimizes actions for a given task. Next, we provide a quick overview of each of these modules.
\subsection{Keypoints as visual latent state and dynamics model}\label{sec:keypt_detector}
 We use an autoencoder with a structural bottleneck to detect 2D keypoints that correspond to pixel positions or areas with maximum variability in the input data. The architecture of the keypoint detector closely follows the implementation in \cite{minderer2019unsupervised}. To train our keypoint detector we collect visual data $\mathcal{D}_\text{key-train}$ for self-supervised keypoint training (see Appendix~\ref{appendix:data_col_det}). 
 After this training phase, we have a keypoint detector that predicts keypoints $z = g_{\text{key}}(o_\text{im})$ of dimensionality $K \times 3$. Here $K$ is the number of keypoints, and each keypoint is given by $z_{k} = ({z}_{k}^{x}, {z}_{k}^{y}, {z}_{k}^{\mu})$, where ${z}_{k}^{x}, {z}_{k}^{y}$ are pixel locations of the $k-\text{th}$ keypoint, and ${z}_{k}^{\mu}$ is its intensity, which corresponds roughly to the probability that that keypoint exists in the image.

Given a trained keypoint detector, we next collect dynamics data to train a dynamics model $\hat{s}_{t+1} = f_\text{dyn}(s_t, u_t)$. The dynamics model is trained to predict the next state, from current state $s_t$ and action $u_t$, where the state $s_t = [z_t, \theta_t]$ combines the low-dimensional visual state $z_t = g_\text{key}(o_{\text{im},t})$ and the joint state $\theta_t$. Actions $u_t$ are desired joint angle displacements. For simple tasks we train this dynamics model on data generated through sine motions on the joints. However for complex tasks, we utilize expert demonstrations to learn this dynamics model. 

\subsection{Gradient-Based Visual MPC towards a keypoint goal state}
We want to optimize an action sequence $\mathbf{u} = (u_0, u_1, \dots, u_T)$ that moves the arm towards the visual goal keypoints $z_\text{goal}$ extracted from a goal image.
Similar to other visual MPC work \cite{ebert2018_visual_foresight, byravan_2018_se3net_control} we utilize our learned visual dynamics model $f_\text{dyn}$ to optimize actions $\mathbf{u}$. Two ingredients are necessary to implement this step: 1) a cost function that measures distances in visual latent space; 2) an action optimizer that can minimize that cost function. We build on the gradient based action optimization presented in \cite{byravan_2018_se3net_control} and extend it for optimizing actions over a time horizon $T$. Specifically, to optimize a sequence of action parameters $\mathbf{u} = (u_0, u_1, \dots, u_T)$ for a horizon of $T$ time steps, we first predict the trajectory $\hat \tau$, that is created through the current $\mathbf{u}$ from starting configuration $s_0$: 
$\hat{s}_1 = f_\text{dyn}(s_0, u_0), 
    ~\hat{s}_2 = f_\text{dyn}(\hat{s}_1, u_1), 
    \dots
    ~\hat{s}_T = f_\text{dyn}(\hat{s}_{t-1}, u_{t-1})$,
which generates a predicted (or planned) trajectory $\hat \tau$.
Intuitively, this step uses the learned dynamics model $f_\text{dyn}$ to simulate forward what would happen if we applied action sequence $\mathbf{u}$. We then measure the cost achieved $C_\psi(\hat \tau, z_\text{goal})$ and perform gradient descent on actions $\mathbf{u}$ such that the cost of the planned trajectory is minimized
\begin{align}
    \mathbf{u}_\text{new} = \mathbf{u} - \eta \nabla_u C_\psi(\hat \tau, z_\text{goal})
\end{align}
Details of our full visual MPC algorithm can be found in the Appendix, in Algorithm 3. Manually designing this cost function is hard, especially in visual feature spaces. In the next section we propose a gradient-based inverse reinforcement learning algorithm to learn this cost function.

%===============================================================================
\section{Gradient-Based IRL from Visual Demonstrations}
\begin{wrapfigure}{R}{0.5\textwidth}
\begin{minipage}{0.5\textwidth}
\vspace{-0.5cm}
\begin{algorithm}[H]
% \vspace{-1cm}
\begin{algorithmic}[1]
\footnotesize{
\STATE{Initial $\psi$, pre-trained $f_\text{dyn}$, learning rates $\eta=.001, \alpha=.01$}
\STATE{ demos $\tau_\text{demo, i}$, with goal state $z_\text{goal} = \tau_T$}
\STATE{initial state $s_0 = (\theta_0, \dot{\theta}_0, z_0)$}
\FOR{each $epoch$}
\STATE{$u_t = 0, \forall t=1,\dots, T$}
\FOR{each $i$ in $\text{iters}_\text{max}$}
\STATE{\com{// rollout $\hat{\tau}$ from initial state $s_0$ and actions $u$}}
\STATE{$\hat{\tau} \gets \text{rollout}(s_0, u, f_\text{dyn})$}
\STATE{\com{// Gradient descent on $u$ with current $C_\psi$}}
\STATE{${u}_{new} \gets {u} - \alpha.\nabla_{{u}}C_{\psi}(\hat{\tau}, z_{goal})$}
\ENDFOR
\STATE{\com{// Update $\psi$ based on ${u_\text{new}}$'s performance}}
\STATE{$\hat{\tau} \gets \text{rollout}(s_0, u_\text{new}, f_\text{dyn})$}
\STATE{\com{// Computes gradient through the inner loop}}
\STATE{$\psi \gets \psi - \eta.\nabla_\psi \irlloss(\hat{\tau}, \tau_\text{demo})$}
\ENDFOR
}
\end{algorithmic}
\caption{\strut\small Gradient-Based IRL for 1 Demo}
\label{algo:cost-learning-meta-train-individual}
\end{algorithm}
\vspace{-0.8cm}
\end{minipage}
\end{wrapfigure}

Most inverse RL algorithms have an inner and outer optimization loop; the inner loop optimizes actions or policies given the current cost function parameters $\psi$, and the outer loop optimizes the cost function parameters given the results of the inner loop. To the best of our knowledge, all existing IRL approaches implement these two optimization steps independently. As we show below, and in our experiments, this can lead to instability in the optimization. Here we derive an algorithm that optimizes cost parameters $\psi$ as \emph{a function} of the inner loop policy optimization step, such that updates to parameters $\psi$ are directly related to their performance in the inner loop.

Specifically, in this work we address deterministic, fixed-horizon and discrete time control tasks with continuous states $\mathbf{s} = (s_1, \dots, s_T)$ and continuous actions $\mathbf{u} = (u_1, . . . , u_T)$. Each state
$s_t = [\theta_t, \dot{\theta}_t, z_t]$ is the concatenation of the measured joint angles and velocities $\theta_t, \dot{\theta}_t$ and the extracted keypoints $z_t$ at time step $t$. The control tasks are characterized by a pre-trained visual dynamics model $\hat{s}_{t+1} = f_\text{dyn}(s_t, u_t)$ and the learned cost function $\learnedCost$.
\subsection{Learning cost functions for action optimization}\label{sec:learning_cost}
In our IRL algorithm, the outer loop optimizes cost parameters $\psi$ and the inner loop optimizes actions $\mathbf{u}$ given the current cost. The result of the inner loop step is a predicted latent trajectory $\hat{\tau}$. Intuitively, we want to learn a cost function $C_\psi$, that, when used in the inner loop, minimizes the IRL loss $\irlloss(\tau_\text{demo}, \hat{\tau})$ between $\hat{\tau}$ and the expert demonstrations $\tau_\text{demo}$. To put it succinctly, we want to compute the gradient of $\irlloss$ wrt to $\psi$: $\nabla_\psi \irlloss$.

To compute this gradient, let's first consider a case where the demonstration consists of only one observation (e.g. the goal) $\tau_\text{demo} = {s_\text{demo}}$, and we want to optimize one action parameter $u$ to achieve this goal in one time step. Then we can write out the IRL optimization problem as

\begin{align}
   \nabla_\psi \irlloss(\tau_\text{demo}, \hat{\tau}_{\psi}) 
   &= \nabla_{\hat{\tau}_{\psi}} \irlloss(\tau_\text{demo}, \hat{\tau}_{\psi}) \nabla_\psi \hat{\tau}_{\psi}\label{eq:chain_rule}\\
   & = \nabla_{\hat{\tau}_{\psi}} \irlloss(\tau_\text{demo}, \hat{\tau}_{\psi}) \nabla_\psi f_\text{dyn}(s, u_\text{opt})\label{eq:dyn_model}\\
   & = \nabla_{\hat{\tau}_{\psi}} \irlloss(\tau_\text{demo}, \hat{\tau}_{\psi}) \nabla_\psi f_\text{dyn}(s, u_\text{init} - \eta \nabla_u C_\psi(s_\text{demo},  f_\text{dyn}(s, u))\label{eq:action_opt}
\end{align}
where in Eq~\ref{eq:chain_rule} we apply the chain rule to decompose $\nabla_\psi \irlloss(\tau_\text{demo}, \hat{\tau}_{\learnedCost})$ into the gradient of $\irlloss$ with respect to the predicted trajectory $\hat{\tau}_\psi$ and the gradient of $\hat{\tau}_\psi$ wrt cost parameters $\psi$. In the next step, Eq~\ref{eq:dyn_model}, we plug in the rollout of the predicted trajectory, which is only one time step, so $\hat{\tau}_\psi=f_\text{dyn}(s, u_\text{opt})$, where $u_\text{opt}$ is the optimized action parameter. In the final step, Eq \ref{eq:action_opt}, we write out the gradient update of the action parameters $u$ which shows the dependence on the cost function $C_\psi$.

This optimization problem is reminiscent of recent gradient-based bi-level optimization approaches to meta-learning \cite{finn2017model, bechtle2019meta}, involving two sets of parameters (in our case $\mathbf{u}, \psi$) to be optimized. Such gradient-based solutions typically involve tracking the gradients through the inner loop, and then auto-differentiating the inner loop optimization trace with respect to the outer parameters. We use the gradient-based optimiser \textit{higher}~\citep{higher} to tackle this bi-level optimization problem. We have described our gradient-based IRL algorithm for inner loops with one step optimization, the extension over multiple time steps requires Eq~\ref{eq:dyn_model} and Eq~\ref{eq:action_opt} to be adapted to the predicted trajectory over $T$ time steps. A high-level overview of our gradient-based IRL algorithm is presented in Algorithm~\ref{algo:cost-learning-meta-train-individual}.

\label{sec:learning_cost_funtions}

\subsection{Cost functions and IRL Loss for learning from visual demonstrations}
\label{sec:cost_function_parametrization}
Our algorithm depends on both, the specification of the IRL loss $\irlloss$ and the cost function parametrization $\learnedCost$.
Intuitively, the $\irlloss$ should measure the distance between the predicted latent trajectory $\hat{\tau}$ and the demonstrated latent trajectory $\tau_\text{demo}$. We would like to keep the $\irlloss$ as simple as possible, and thus choose it to be the squared distance between predicted and demonstrated keypoints at each time step, $\irlloss(\tau_\text{demo}, \hat{\tau}) = \sum_t (z_{t, \text{demo}} - \hat{z}_t)^2$. Similar to \cite{englert2017inverseRL}, we compare three distinct parametrizations for the cost function $C_\psi$:
 
\textbf{Weighted Cost} $C_\psi(\hat{\tau}, z_\text{goal}) =  \sum_k \left[\psi^{x}_{k} \sum_t (\hat{z}_{t, k}^{x} - z_{\text{goal}, k}^{x})^2 + \psi^{y}_{k} \sum_t (\hat{z}_{t, k}^{y} - z_{\text{goal}, k}^{y})^2 \right]
    % \vspace{-0.5cm}
$

where $\hat{z}_{t, k}^{x} , \hat{z}_{t, k}^{y}$ is the $k$th predicted keypoint at time-step $t$ and $z_\text{goal, k}^{x}, z_\text{goal, k}^{y}$ is the goal keypoint. This simple cost function parametrization provides a constant weight per $x, y$ dimension of each key point. This cost function has $K \times 2$ parameters.
    
\textbf{Time Dependent Weighted Cost} $C_\psi({\hat{\tau}}, z_\text{goal}) =  \sum_k \sum_t \left[ \psi^{x}_{t,k} (\hat{z}_{t, k}^{x} - z_{\text{goal}, k}^{x})^2 + \psi^{y}_{t,k}  (\hat{z}_{t, k}^{y} - z_{\text{goal}, k}^{y})^2 \right]
$

This cost extends the previous formulation to provide a weight for each time step $t$. This adds more flexibility to the cost and allows to capture time-dependent importance of specific keypoints. This cost function has $T \times K \times 2$ parameters, which scale linearly with the horizon length.

\textbf{RBF Weighted Cost} $C_\psi(\hat{\tau}, {z}_\text{goal}) =  \sum_k \sum_t \sum_j \left[ \psi^{x}_{j,k}(t) (\hat{z}_{t, k}^{x} - z_{\text{goal}, k}^{x})^2 + \psi^{y}_{j,k}(t)  (\hat{z}_{t, k}^{y} - z_{\text{goal}, k}^{y})^2 \right]
$

Here we introduce $J$ time dependent RBF kernels $\psi_{j,k}(t) = \exp{(b (t - \mu_j)^2)}$. This cost allows us to more easily scale to longer time horizons, with $J \times K \times 2$ parameters, and $J < T$. Kernels are uniformly spaced in time and $b$ is chosen to create some overlap between neighboring kernels.

\subsection{Illustrative Comparison with Feature-Matching IRL}\label{sec:simulation_experiment}

\begin{wrapfigure}{l}{0.55\textwidth}
    \begin{subfigure}{0.27\textwidth}
        \includegraphics[width=1.0\textwidth]{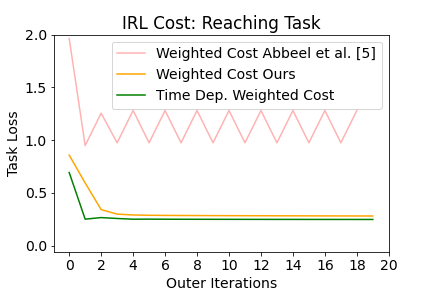}
        \includegraphics[width=1.0\textwidth]{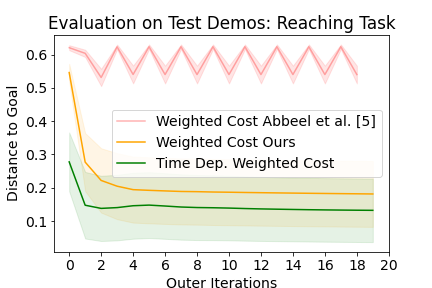}
        \subcaption{}
    \end{subfigure}
    \begin{subfigure}{0.27\textwidth}
        \includegraphics[width=1.0\textwidth]{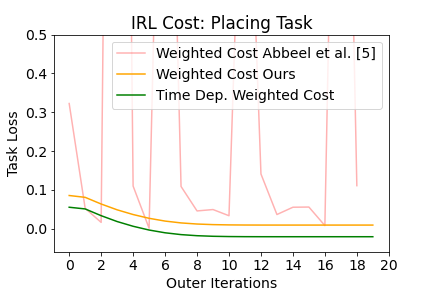}
        \includegraphics[width=1.0\textwidth]{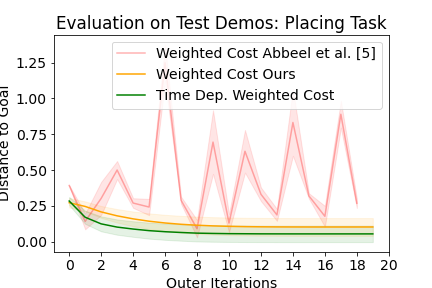}
        \subcaption{}
    \end{subfigure}
    \caption{\small (Top) IRL cost during cost training for reaching (a) and placing (b) task with one demonstration. (Bottom) Performance of learned cost on five test tasks. We compare our learned IRL costs with a cost trained using apprenticeship learning~\citep{abbeel2004apprenticeship} .}
    \vspace{-0.3cm}
\label{fig:sim_experiment}
\end{wrapfigure}
Here, we illustrate the differences between our approach and feature matching IRL approaches in terms of optimization behavior on simulation tasks with known models. We compare our method to the IRL apprenticeship learning algorithm from \citep{abbeel2004apprenticeship} for a reaching and a placing task on a simulated Kuka robot. We adapted the code from \citep{Apprenticeship_Code} for our experiments (see appendix \ref{appendix:apprenticeship_algo}). We assume to be given ground truth keypoints in 3-D, placed on an object that the Kuka holds. Furthermore we use a differentiable model that predicts keypoint changes for applied actions.

We train a weighted cost using only one demonstration, and evaluate the learned cost function on five test demonstrations for with our algorithm and our baseline. In Figure \ref{fig:sim_experiment} we show convergence on training and test tasks, as a function of outer loop iterations. We see that our baseline oscillates between good and bad solutions, while our algorithm converges to a good solution.  
We believe this improvement in convergence behavior is due to the presence of an explicit connection between the policy optimization and the cost function parameter learning in our method. This connection allows us to compute gradients that communicate between inner and outer loops and thus explicitly account for the cost function performance for policy optimization during cost function learning. In contrast, existing model-based IRL approaches, such as the feature matching algorithm, separate the outer and inner loop and rely on careful design of multiple constraints or features to update cost function parameters.% \cite{englert2017inverseRL}.

\section{Hardware Experiments}

\label{sec:hardware_exps}
\begin{wrapfigure}{R}{0.25\textwidth}
\vspace{-0.5 cm}
\begin{minipage}{0.25\textwidth}
\begin{figure}[H]
    \centering
    \vspace{-0.5 cm}
    \includegraphics[width=.45\textwidth]{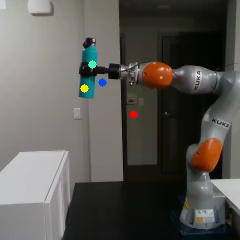}
    \includegraphics[width=0.45\textwidth]{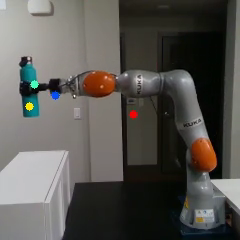}
    \small \caption{\small Reaching task} 
    \label{fig:hardware_tr_demos}
        \vspace{-0.4cm}
\end{figure}
\end{minipage}
\end{wrapfigure}
We evaluate the proposed approach for inverse reinforcement learning from visual demonstrations by performing a sequence of qualitative and quantitative experiments. We seek to interpret the learned cost functions and investigate their ability to successfully reproduce the demonstrated tasks. In our experiments, we assume we have pre-trained a key point detector (see Figure~\ref{fig:hardware_tr_demos}), and a good enough visual dynamics model to accomplish the task. We use the same keypoint detector for all experiments. Details about training the keypoint detector and the dynamics model can be found in the Appendix.
\subsection{Quantitative Analysis on automatically generated visual demonstrations}\label{experiment_1_data_col}

We collect a set of $15$ automatically generated demonstrations of moving an object from one (visual) location to another using the KUKA arm, making sure that visually the object moves only in the X-axis (see Figure \ref{fig:hardware_parameter_vis_results} for an example). Constraining the movement of the gripped object in this way allows us to interpret the learned cost functions better. We also note that one of our 4 keypoints (in red), is fixed in the background. The collected demonstrations comprise the start state $\theta_0, \dot{\theta}_0$, and keypoint observations $z_t = g_\text{key}(o_\text{im,t})$, for $T=25$ frames at a frame rate of $5Hz$. The keypoint detector predicts $4$ keypoints per frame. We train the parametrized costs described in Section \ref{sec:cost_function_parametrization} with $1$ and $10$ reaching demonstrations; and evaluate their performance by optimizing an action policy using the learned costs on $5$ test demonstrations.

We compare our IRL algorithm to 2 baselines: (1) the IRL apprenticeship learning algorithm \cite{abbeel2004apprenticeship} combined with the weighted cost from \ref{sec:cost_function_parametrization}, and (2) %no IRL, and instead use
a naive (``Default") cost that measures the distance between the predicted and goal keypoint. This cost is defined as $C_\text{default} = \sum_t^T (\hat z_t - z_\text{goal})^2$ for a trajectory with $T$ steps. For visual model-predictive control via learned (or default) cost, the learning rate for action optimization is chosen to be the same as during the IRL training phase, $\eta=0.001$.
\subsubsection{Training and Analysis of the Cost Functions} \label{sec:experiment_1_training_analysis}
\begin{figure}[H]
    \centering
    
    {\begin{subfigure}[b]{0.23\textwidth}
    \includegraphics[width=1.0\textwidth]{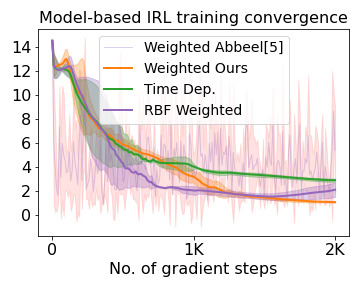}
    \subcaption{1 demo, train}
    \end{subfigure}}
    {\begin{subfigure}[b]{0.26\textwidth}
    \includegraphics[width=1.0\textwidth]{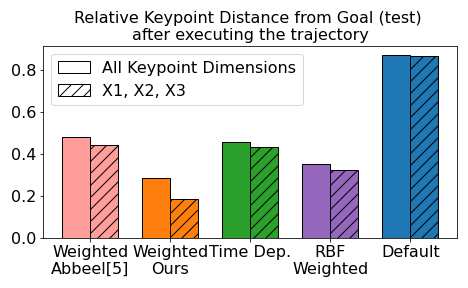}
    \subcaption{1 demo, test}
    \end{subfigure}}
    {\begin{subfigure}[b]{0.23\textwidth}
    \includegraphics[width=1.0\textwidth]{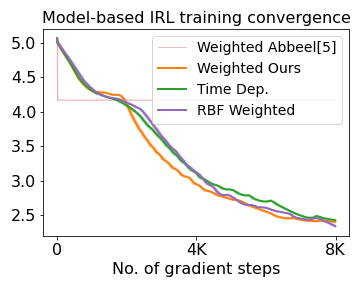}
    \subcaption{10 demos, train}
    \end{subfigure}} 
    {\begin{subfigure}[b]{0.26\textwidth}
    \includegraphics[width=1.0\textwidth]{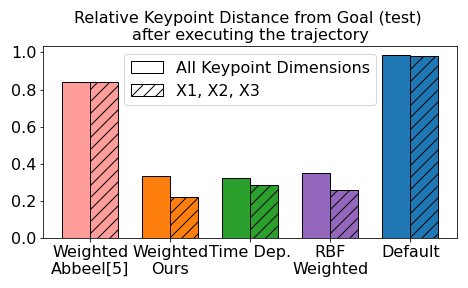}
    \subcaption{10 demos, test}
    \end{subfigure}}

    \caption{\small \textbf{IRL training and test evaluation} (a) and (c) show the $\irlloss$ during training of the parametrized costs from 1 and $\nsincosdemo$ demos. Figures (b, d) show the relative distance to the goal keypoint achieved at test time when optimizing the action trajectory with the learned costs and baselines. Results are averaged across 3 seeds.
    } 
    \label{fig:hardware_cost_training_results}
    \vspace{-0.5cm}
\end{figure}

Figure~\ref{fig:hardware_cost_training_results} depicts the results achieved on the simple reaching task. The final relative distance (see Appendix) to goal keypoint positions from the planned trajectory is considerably less when optimized using all three of the learned costs compared to both baselines (see (b, d)). We calculate and compare this metric for all keypoint dimensions as well as only the dimensions corresponding to $z^{x}_{t, 1}, z^{x}_{t, 2}$ and $z^{x}_{t, 3}$, which are the least noisy keypoint observation dimensions. We also note that the learned costs perform overall similarly irrespective of whether they were trained on a single demonstration or on ten demonstrations. This observation encouraged the use of a single demonstration for the next set of experiments (Section \ref{sec:human_demos}), where such demonstrations are harder to acquire. 

\begin{figure}[t]
% \vspace{-0.5cm}
    \centering
    \begin{subfigure}[b]{0.245\textwidth}
    \includegraphics[width=1.0\textwidth]{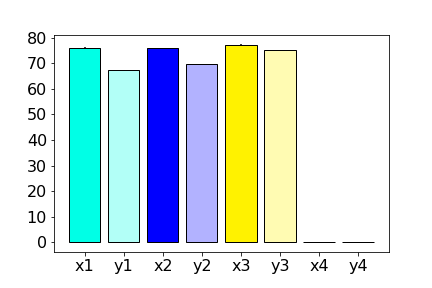}
    % \subcaption{Mean IRL Cost for training data \\ while Cost Function Learning }
    \subcaption{Weighted (ours)}
    \end{subfigure}
    \begin{subfigure}[b]{0.245\textwidth}
    \includegraphics[width=1.0\textwidth]{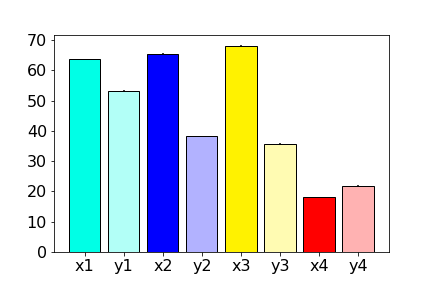}
    % \subcaption{Mean IRL Cost for training data \\ while Cost Function Learning }
    \subcaption{TimeDep (ours)}
    \end{subfigure}
    \begin{subfigure}[b]{0.245\textwidth}
    \includegraphics[width=1.0\textwidth]{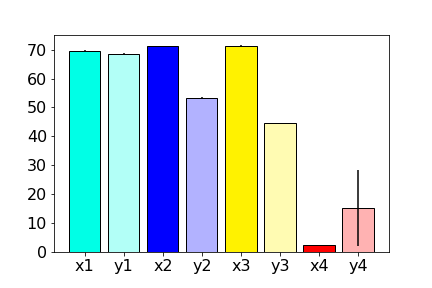}
    % \subcaption{Mean IRL Cost for training data \\ while Cost Function Learning }
    \subcaption{RBF Weighted (ours)}
    \end{subfigure}
    \begin{subfigure}[b]{0.245\textwidth}
    \includegraphics[width=1.0\textwidth]{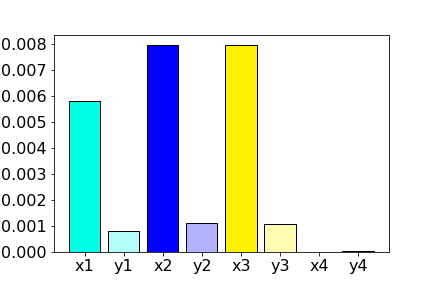}
    % \subcaption{Mean IRL Cost for training data \\ while Cost Function Learning }
    \subcaption{Weighted (baseline)}
    \end{subfigure}
    \caption{\small \textbf{Learned cost Parameters:} corresponding to the keypoint vector's dimensions after training on $\nsincosdemo$ demos. $Y$-axes and one keypoint (in red) receive less weight. Colors are matched to keypoints shown in Fig \ref{fig:hardware_tr_demos}.
    } 
    \label{fig:hardware_parameter_vis_results}
    %\vspace{-0.5cm}
\end{figure}

As noted before, the $10$ reaching demonstrations we used for training had very little variability for the visual keypoints along the $Y$-direction and for one particular keypoint (marked red in Figure \ref{fig:hardware_tr_demos}). Figure \ref{fig:hardware_parameter_vis_results} (a,b,c) illustrate that all of the proposed parametrized cost functions learn relatively small weights corresponding to the $Y$-axes and the red keypoint. 
indicating that they have identified properties of the visual demonstrations they have been trained on. 
Finally, the parameters of baseline(1) (while being significantly smaller) have a similar weight structure to the rest of the models. This indicates that they are able to capture the demonstration properties. However, their overall performance during evaluation was far worse than our learned costs due to the lesser weight each parameter bears. Note that we could tune the learning rate $\eta$ with which actions are optimized at test time to account for these smaller weights, which would improve performance of our baseline. However, this is not necessary for our algorithm, which learned to scale cost function parameters wrt to the $\eta$ used during the IRL training phase. Furthermore, the IRL optimization procedure for baseline (1) was very unstable, and it was unclear whether the algorithm has or will converge. Previous work has proposed to scale and regularise the learned weights as done in the maximum entropy literature~\citep{kalakrishnan_2013_irl, boularias2011relativeirl} or define additional constraints~\citep{englert2017inverseRL} to address some of these issues. We believe one reason for this instability is that the inner and outer loops are disconnected in such feature matching algorithms. Our algorithm instead connects the inner and outer loop optimization steps, and is therefore able to leverage gradient updates from action optimization in the inner loop for learning cost function parameters that automatically work well on the desired task without any additional help.
\subsection{Learning Cost Functions from Visual Human Demonstrations} \label{sec:human_demos}
In this subsection we scale the proposed method to a more challenging task both from manipulation and demonstration points of view. We consider the task of placing a  bottle on a shelf demonstrated by a human user through video data.
%\vspace{-0.3cm}
\paragraph{Expert Demonstration Data Collection}
We collect the human demonstration at a frame rate of $30$ Hz, which we then downsample to $5$ Hz. In contrast to Section \ref{experiment_1_data_col}, we do not have access to the initial proprioceptive state $\theta, \dot{\theta}$. 
We therefore test with 2 starting configurations of the robot. Start configuration 1) we choose an initial position for the robot that is roughly close to the human demonstration's initial position; and start configuration 2) that is closer to the target. We preprocess all the video-frames to obtain keypoint vectors corresponding to each step, relative to the first frame.
%

%\vspace{-0.3cm}
\paragraph{Training the Cost Functions and analysis}
We experiment on a task that is comprised of two individual motions. During the first half of the demonstration, the object moves only along the X-axis towards the shelf, while in the second half it moves downwards (i.e. along the Y-axis, while X-coordinate of the object remains constant). We train the $3$ cost function architectures from Section \ref{sec:cost_function_parametrization} on a single human demonstration for placing a bottle for $5000$ gradient steps. 
\begin{figure}[H]
    \centering
	\begin{subfigure}[b]{0.23\textwidth}
    \includegraphics[width=1.0\textwidth]{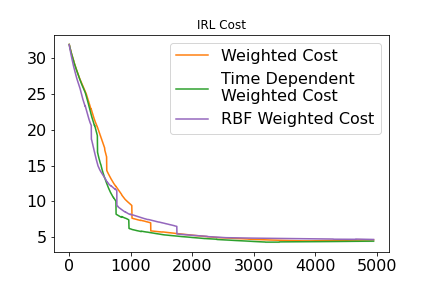}\label{human_demo_eval_irl}
    \subcaption{Training: IRL Loss}
    \end{subfigure}
    \begin{subfigure}[b]{0.23\textwidth}
    \includegraphics[width=1.0\textwidth]{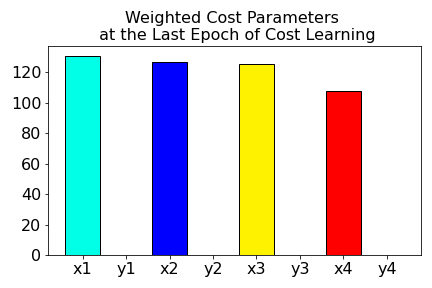}
    \subcaption{Weighted}
    \end{subfigure}
    \begin{subfigure}[b]{0.23\textwidth}
    \includegraphics[width=1.0\textwidth]{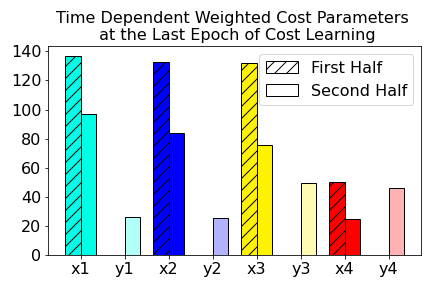}
    \subcaption{TimeDep Weighted}
    \end{subfigure}
    \begin{subfigure}[b]{0.23\textwidth}
    \includegraphics[width=1.0\textwidth]{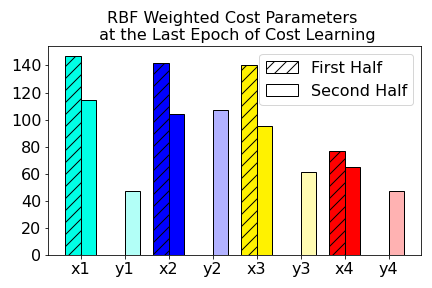}
    \subcaption{RBF Weighted}
    \end{subfigure}
    
    \caption{\small (a) plots the $\irlloss$  while training costs for 5K gradient steps with a human demonstration. (b, c and d) show the values of the learned costs' parameters. For Time Dependent (b) and RBF costs with 5 kernels (c) which calculate separate parameters corresponding to each step of the trajectory, we compare the mean of the parameters corresponding to each keypoint across the first five steps to the last five steps.
    } 
    \label{fig:human_demo_parameter_vis_results}
    \vspace{-0.5cm}
\end{figure}
The $\irlloss$ loss converges roughly around $2K$ iterations (Figure \ref{fig:human_demo_parameter_vis_results} (a)). We note that the parameters of the time-dependent cost functions (Figure \ref{fig:human_demo_parameter_vis_results} (b and c)) learn to emphasize the distance from the goal in the X direction during the first half of the motion and Y-direction in the latter half.
\subsubsection{Using the Learned Cost Function on the Robot}
\begin{figure}[H]
    \centering
    \begin{subfigure}[b]{0.14\textwidth}
    \includegraphics[width=1.0\textwidth]{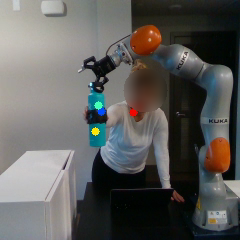}
    %\subcaption{}
    \end{subfigure}
    \begin{subfigure}[b]{0.14\textwidth}
    \includegraphics[width=1.0\textwidth]{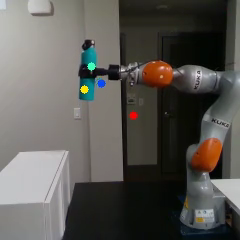}
    %\subcaption{}
    \end{subfigure}
    \begin{subfigure}[b]{0.14\textwidth}
    \includegraphics[width=1.0\textwidth]{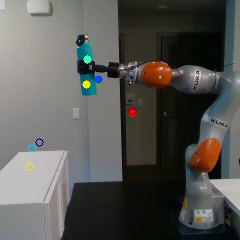}
    %\subcaption{}
    \end{subfigure}
    \begin{subfigure}[b]{0.14\textwidth}
    \includegraphics[width=1.0\textwidth]{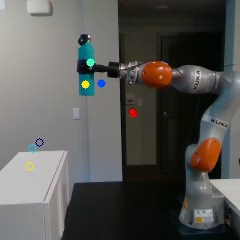}
    %\subcaption{}
    \end{subfigure}
    \begin{subfigure}[b]{0.14\textwidth}
    \includegraphics[width=1.0\textwidth]{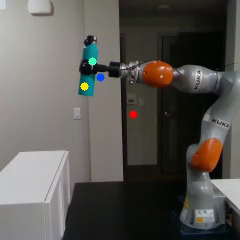}
    %\subcaption{}
    \end{subfigure}\\
    \begin{subfigure}[b]{0.14\textwidth}
    \includegraphics[width=1.0\textwidth]{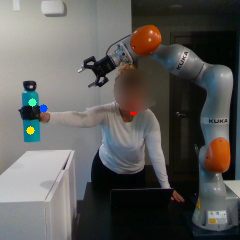}
    %\subcaption{}
    \end{subfigure}
    \begin{subfigure}[b]{0.14\textwidth}
    \includegraphics[width=1.0\textwidth]{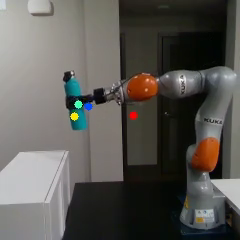}
    %\subcaption{}
    \end{subfigure}
    \begin{subfigure}[b]{0.14\textwidth}
    \includegraphics[width=1.0\textwidth]{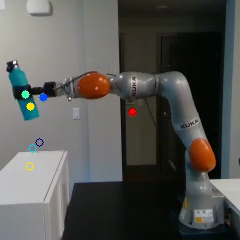}
    %\subcaption{}
    \end{subfigure}
    \begin{subfigure}[b]{0.14\textwidth}
    \includegraphics[width=1.0\textwidth]{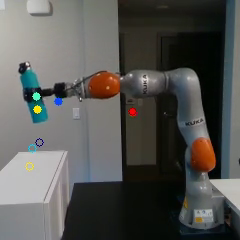}
    %\subcaption{}
    \end{subfigure}
     \begin{subfigure}[b]{0.14\textwidth}
    \includegraphics[width=1.0\textwidth]{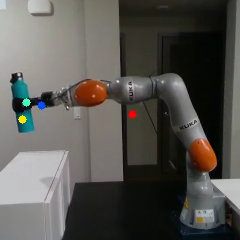}
    %\subcaption{}
    \end{subfigure}\\
    \begin{subfigure}[b]{0.14\textwidth}
    \includegraphics[width=1.0\textwidth]{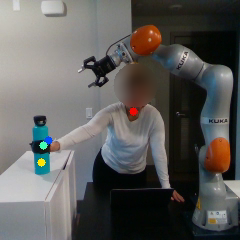}
    \subcaption{\small{Demo}}
    \end{subfigure}
    \begin{subfigure}[b]{0.14\textwidth}
    \includegraphics[width=1.0\textwidth]{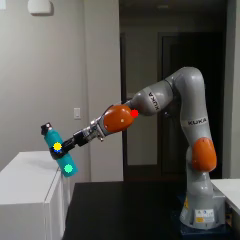}
    \subcaption{\small{Default}}
    \end{subfigure}
    \begin{subfigure}[b]{0.14\textwidth}
    \includegraphics[width=1.0\textwidth]{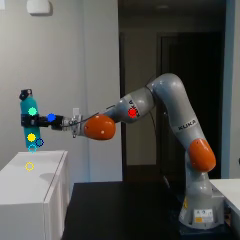}
    \subcaption{\small{Weighted}}
    \end{subfigure}
    \begin{subfigure}[b]{0.14\textwidth}
    \includegraphics[width=1.0\textwidth]{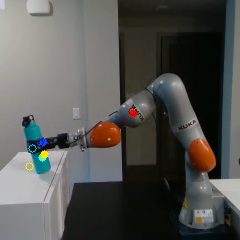}
    \subcaption{\small{TimeDep}}
    \end{subfigure}
    \begin{subfigure}[b]{0.14\textwidth}
    \includegraphics[width=1.0\textwidth]{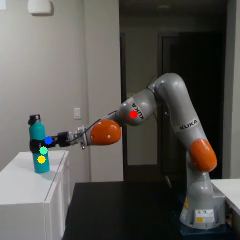}
    \subcaption{\small{RBF}}
    \end{subfigure}
\caption{
\small{Column a) Human Demonstration that is used for the IRL algorithm to extract cost functions. Column b)-d) Comparison of visual MPC result using the default and learned costs. First row corresponds to timestep $t=0$, middle row to $t=5$ and bottom row to $t=10$ of executing the placing task. The detected and goal keypoints in each image are depicted using filled and hollow circles respectively.}
}
\label{fig:placing_task}
\vspace{-0.5cm}
\end{figure}
We use the $3$ learned cost functions and our pre-trained visual dynamics model to optimize a sequence of $T=10$ desired joint angle displacements towards the keypoint goal from demonstraion. We record the mean squared distance to the goal keypoint in Table \ref{fig:human_demo_eval}. We note that while both Time Dependent and the RBF Weighted Costs perform much better than our baseline, the simple Weighted Cost performs well on just one of the test cases, indicating that the time-dependency component of the cost leads to better generalization.
\begin{table}[H]
\centering
		\begin{footnotesize}
% 		\vspace{1mm}
			\begin{tabular}[b]{p{0.5 cm} c c c c c c c c}
				\toprule 
				\bf Start & \bf Weighted & \bf TimeDep & \bf RBF& \bf Default\\ 
				& Mean (Std) & Mean (Std) & Mean (Std) & Mean (Std) \\
				\midrule
				1 & 40.99 (9.08) & 6.12 (0.94) & \textbf{4.26} (\textbf{1.20}) & 26.96 (5.41) \\
                2 & 3.61 (0.40) & \textbf{3.53} (\textbf{0.21}) & 4.40 (0.15) & 15.76 (1.34) \\
				\bottomrule
				% \subcaption{}
			\end{tabular}
		\end{footnotesize}
		\vspace{0.1cm}
		\caption{\label{fig:human_demo_eval} \small records the mean squared distance between the keypoints obtained after executing an action trajectory optimized from the indicated cost on the KUKA to the given goal keypoints from 2 starting configurations.}
		\vspace{-0.7cm}
\end{table}
%===============================================================================
\section{Discussion and Future Work}
\label{sec:conclusion}
We propose a gradient-based IRL framework that learns cost function from visual human demonstrations. We learn a compact keypoint-based image representation, and train a visual dynamics in that latent space. We then use the keypoint trajectories extracted from user demonstrations, and our learned dynamics model, to learn different cost functions using our gradient-based IRL algorithm.

Several challenges remain: Learning a good visual predictive model is difficult, and created one of the main challenges in this work. One avenue for easier dynamics model training is to robustify the keypoint detector using methods like \citet{florence2018dense}, so that it becomes invariant to different viewpoints.
Furthermore, our work assumes that demonstrations are given from the perspective of the robot. We account for different starting configurations by learning on relative demonstrations instead of absolute. A step towards generalizing our approach even more is  to consider methods that can map demonstrations from one context to another, as was presented in \citet{liu2018imitation}. Finally, while we have presented experimental results for the more improved convergence behavior of our gradient-based IRL algorithm, as compared to the feature-matching baseline, we would like to investigate our findings in more depth in future work.

%===============================================================================

% The maximum paper length is 8 pages excluding references and acknowledgements, and 10 pages including references and acknowledgements

\clearpage
% The acknowledgments are automatically included only in the final version of the paper.
\acknowledgments{We would like to thank Kristen Morse for her useful suggestions during the preparation of this manuscript as well as Deepak Pathak and Masoumeh Aminzadeh for discussions during the early stages of this project.}

%===============================================================================
\bibliography{bibliography} 
\clearpage
\appendix
\include{appendix_corl_final_draft}

\end{document}

%% file: appendix_corl_final_draft.tex
\section{Appendix}
Our framework consists of several components trained in isolation, eg the keypoint detector and the dynamics model. Here we describe the architecture and training details of both. Furthermore, we go into the details of our baseline implementation (A.4) as well as our visual MPC trajectory optimization step (A.5). Finally we also visually depict the results for the 2nd starting configuration of experiments done in Section 5.2 (A.6).
\subsection{Keypoint Detector And Dynamics Model Architectures}
Here we describe the architecture details of both the keypoint detector and the dynamics model.
\paragraph{keypoint detector $g_\text{key}$:} The complete architecture for training the keypoint detector comprises of an autoencoder with a structural bottleneck that can extract "significant" 2D locations from the input images. $g_{key}$ itself is essentially the encoder component of the autoencoder. Following \citet{DIGIT}, we implement $g_{key}$ as a mini version of ResNet-18. The input images used are cropped to a resolution of $[240 \times 240]$.

\paragraph{dynamics model $f_{\text{dyn}}$:} Our dynamics model $\hat{s}_{t} = f_\text{dyn}(\hat s_{t-1}, u_{t-1})$, where $\hat{s}_t = [\hat z_t, \hat \theta_t, \hat{\dot{\theta}}_t]$, has 2 components:

1) a keypoint predictor $f_{\text{mlp}}$ 
        \begin{equation*}
            \hat z_t = f_{\text{mlp}}(\hat s_{t-1}, u_{t-1});
        \end{equation*} 
which is modeled by a neural network with two hidden layers with $100$ and $25$ neurons respectively and a ReLu activations after each layer except the last.

2) a next joint state predictor which simply integrates the action $u_{t-1}$, which are desired joint angle displacements, with the current (predicted) joint state $\hat \theta_{t-1}, \hat{\dot{\theta}}_{t-1}$ to predict the next state:
        \begin{align*}
            \hat \theta_t &= \hat \theta_{t-1} + u_{t-1}\\
            \hat{\dot{\theta}}_t &= \hat{\dot{\theta}}_{t-1}
        \end{align*}

\subsection{Self-Supervised Training of Keypoint detector}\label{appendix:data_col_det}
To train our keypoint detector we collect 108 sequences of video data, each 10 frames long. For each sequence we move the the Kuka iiwa while gripping an object into a random configuration, and then only move the last joint such that the detector emphasizes on extracting 2D locations that correspond to the gripped object as opposed to the robot arm. We train the keypoint detector until convergence. For additional details regarding the training process refer to \citet{minderer2019unsupervised}. The resulting keypoint detector is visualized in Figure \ref{fig:hardware_tr_demos}.
\subsection{Training of Dynamics Model in Latent Space}
\label{appendix:data_col_dyn}
We train 2 separate dynamics models $f_\text{dyn}$ for the two sets of experiments. 

\paragraph{Experiments of Section 5.1} For the first task of moving the bottle in 'x'-direction we use a purely self-supervised data collection routine. We command sine motions at various frequencies and amplitudes to each joint of the arm. The sine motions were designed to move joints 2,4 and 6 the most, such that the arm stays in plane. The trained keypoint detector is running asynchronously at $30hz$, and outputs $z_t$ at that rate. We collect tuples $(s_{t}, u_t, s_{t+1})$, where $s_t = [\theta_t, \dot \theta_t, z_t]$, at a frequency of $5$Hz, on which we train the dynamics model.

The parameters of $f_{\text{dyn}}$ are trained by optimizing a normalized mean squared error (NMSE) between predicted $\hat s_{t+1} = f_\text{dyn}(s_t, u_t)$ and the ground truth $s_{t+1}$. We train this model until we converge to a NMSE of $0.3$.

\paragraph{Experiments of Section 5.2} For the task of placing a bottle, we combine self-supervised data collection, with data collected from expert controllers (that roughly achieve the placing task), and data augmentation techniques. The self-supervised data collection is similar to the one described above. The trained keypoint detector is running asynchronously at $30hz$, and outputs $z_t$ at that rate. We collect tuples $(s_{t}, u_t, s_{t+1})$, where $s_t = [\theta_t, \dot \theta_t, z_t]$, at a frequency of $1$Hz, to train the dynamics model.

The parameters of $f_{\text{dyn}}$ are trained by optimizing a normalized mean squared error (NMSE) between predicted $\hat s_{t+1} = f_\text{dyn}(s_t, u_t)$ and the ground truth $s_{t+1}$. We were able to train this model to a NMSE of $0.03$.

\subsection{Adaptation of Abeel's IRL algorithm}
\label{appendix:apprenticeship_algo}
We extend the publicly available implementation~\citep{Apprenticeship_Code} for our baseline comparison of \citet{abbeel2004apprenticeship}. We change their inner loop, to use our model-based trajectory optimisation routine. Further, we saw fair to use as features $\phi(\cdot)$ the per-step task objective $\irlloss(\cdot)$ employed in this work.  Finally, ~\citep{Apprenticeship_Code} had a larger value for the minimal distance from baseline constraint (2 instead of 1) than the one suggested in the original paper~\citep{abbeel2004apprenticeship}. We found that using 1 as advised by the authors to work better than~\citep{Apprenticeship_Code}. Therefore, the overall algorithm remains similar to the introduced by~\citet{abbeel2004apprenticeship} and namely, 
\begin{algorithm}[H]
% \vspace{-1cm}
\begin{algorithmic}[1]
\footnotesize{
\STATE{Randomly initialise parameters $u(0)$ and pre-trained $f_{dyn}$, compute $\mu(0)=\mu(\pi(0))$.}
\STATE{demo $\tau_\text{demo}$, with goal state $z_\text{goal} = \tau_{T}$}
\STATE{initial state $s_0 = (\theta_0, \dot{\theta}_0, z_0)$}
\FOR{each $epoch$}
\STATE{$u_t=0, \forall t=1,\dots, T$}
\vspace{2mm}
\STATE{\com{$//$ Take maximal $\psi$ by using the expectation of features from the final rollouts.}}
\STATE{$\irlloss = max_{\psi:||\psi||_{2}\leq1} min_{j\in[0..(epoch-1)]} \psi^{T}(\mu_{E} - \mu(j))$}
\STATE{let $\psi(epoch)$ be the value of $\psi$ that attains this max.}
\FOR{each $i$ in $\text{iters}_\text{max}$}
\vspace{2mm}
\STATE{\com{$//$ rollout $\hat{\tau}$ from initial state $s_0$ and actions $u$}}
\STATE{$\hat{\tau} \gets \text{rollout}(s_0, u, f_\text{dyn})$}
\vspace{2mm}
\STATE{\com{$//$ Gradient descent on $u$ with current $C_{\psi}=\psi^T \phi(\cdot)$}}
\STATE{${u}_{new} \gets {u} - \alpha.\nabla_{{u}}C_{\psi}(\hat{\tau}, \psi)$}
\ENDFOR
\vspace{2mm}
\STATE{\com{$//$ Current expectation of features $\mu(\cdot)$ is over all features from the final rollout.}}
\STATE{Compute $\mu(\cdot) = \mathbb{E}[\sum_{t=0}^{T}\gamma^{t}\phi(z_t)]$.}
\ENDFOR
}
\end{algorithmic}
\caption{\strut\small Apprenticeship Learning Algorithm}
\label{algo:irl-apprenticeship}
\end{algorithm}

\subsection{Details regarding the Evaluation of different cost functions}\label{appendix:relative_dist}

We evaluate the baseline and learned cost functions by comparing the keypoint from the last step of planned trajectory they optimize with the goal keypoint. The planned trajectory is extracted using Algorithm \ref{algo:planning}. Our evaluation metric \textit{relative distance} is defined as $\frac{|| {\tilde z}_{T} - {z}_{goal}||}{||{\tilde z}_{0} - {z}_{goal}||}$

\begin{algorithm}[H]
% \vspace{-1cm}
\begin{algorithmic}[1]
\footnotesize{
\STATE{Given the cost $C_\psi$, planning horizon $T$, the forward dynamics model $f_\text{dyn}$ and the initial state $s_0 = [{z}_0, \theta_0]$ .... where ${z}_t$, $\theta_t$ denote the keypoint and joint vector at time $t$ and ${z}_\text{goal}$ denotes the goal keypoint vector.}
\vspace{2mm}
\STATE{Initialize ${u}_{\text{init}, t} = 0, \forall t=1, \dots, T$}
\vspace{2mm}
\STATE{\com{$//$ Rollout using the initial actions}}
\STATE{$\hat z_0 = z_0, \hat \theta_0 = \theta_0$}
\STATE{$\hat \tau = \{\hat z_0\}$}
\FOR{\hspace{1mm}$t \gets 1:T$}
\STATE{$\hat s_{t-1} = [\hat z_{t=1}, \hat \theta_{t-1}, \hat{\dot{\theta}}_{t-1}]^T$}
\STATE{$\hat z_t, \hat \theta_t, \hat{\dot{\theta}}_{t} = f_\text{dyn}(\hat s_{t-1}, u_{\text{init}, t-1})$}
\STATE{$\hat \tau \gets \hat \tau \cup \hat{z}_t$}
\ENDFOR

\vspace{2mm}
\STATE{\com{$//$Action optimization}}
\STATE{${u}_\text{opt} \gets {u_\text{init}} - \alpha.\nabla_{{u}}C(\hat \tau, z_{goal})$}

\vspace{2mm}
\STATE{\com{$//$Get planned trajectory by rolling out ${u}_\text{opt}$}}
\STATE{$\tilde z_0 = z_0, \tilde \theta_0 = \theta_0, \tilde{\dot{\theta}}_0 = \dot{\theta}_0$}
\FOR{$t \gets 1:T$}
\STATE{$\tilde z_t, \tilde \theta_t = f_\text{dyn}([\tilde z_{t-1}, \tilde \theta_{t-1}, \tilde{\dot{\theta}}_{t-1}], u_{\text{opt}, t-1})$}
\ENDFOR
}
\STATE{Return $\tilde z, \tilde \theta$}
\end{algorithmic}
\caption{\strut\small Trajectory planning using given Cost}
\label{algo:planning}
\end{algorithm}

\subsection{Additional Results}
\begin{figure}[H]
    \centering
    \begin{subfigure}[b]{0.15\textwidth}
    \includegraphics[width=1.0\textwidth]{hardware_exps/placing-robot-defaultcost-0.png}
    \subcaption{Test Case 1}
    \end{subfigure}
    \begin{subfigure}[b]{0.15\textwidth}
    \includegraphics[width=1.0\textwidth]{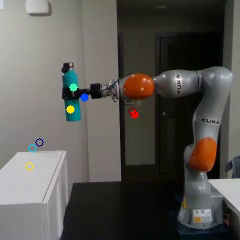}
    \subcaption{Test Case 2}
    \end{subfigure}
    \caption{The 2 starting configurations for the placing task we evaluate our approach on.}
    \label{fig:placing-task-test-cases}
\end{figure}

\begin{figure}[H]
    \centering
    \begin{subfigure}[b]{0.15\textwidth}
    \includegraphics[width=1.0\textwidth]{hardware_exps/placing-demo-0.png}
    %\subcaption{}
    \end{subfigure}
    \begin{subfigure}[b]{0.15\textwidth}
    \includegraphics[width=1.0\textwidth]{hardware_exps/TrimmedDefault2_start.png}
    %\subcaption{}
    \end{subfigure}
    \begin{subfigure}[b]{0.15\textwidth}
    \includegraphics[width=1.0\textwidth]{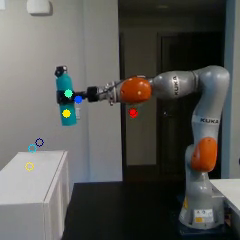}
    %\subcaption{}
    \end{subfigure}
    \begin{subfigure}[b]{0.15\textwidth}
    \includegraphics[width=1.0\textwidth]{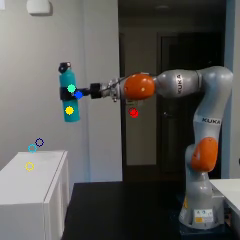}
    %\subcaption{}
    \end{subfigure}
    \begin{subfigure}[b]{0.15\textwidth}
    \includegraphics[width=1.0\textwidth]{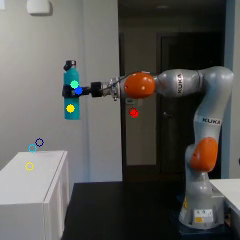}
    %\subcaption{}
    \end{subfigure}\\
    \begin{subfigure}[b]{0.15\textwidth}
    \includegraphics[width=1.0\textwidth]{hardware_exps/placing-demo-1.png}
    %\subcaption{}
    \end{subfigure}
    \begin{subfigure}[b]{0.15\textwidth}
    \includegraphics[width=1.0\textwidth]{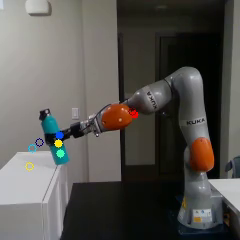}
    %\subcaption{}
    \end{subfigure}
    \begin{subfigure}[b]{0.15\textwidth}
    \includegraphics[width=1.0\textwidth]{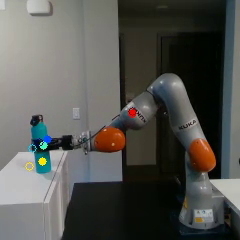}
    %\subcaption{}
    \end{subfigure}
    \begin{subfigure}[b]{0.15\textwidth}
    \includegraphics[width=1.0\textwidth]{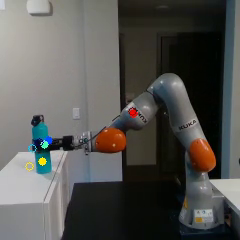}
    %\subcaption{}
    \end{subfigure}
     \begin{subfigure}[b]{0.15\textwidth}
    \includegraphics[width=1.0\textwidth]{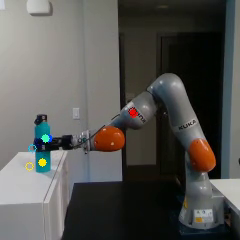}
    %\subcaption{}
    \end{subfigure}\\
    \begin{subfigure}[b]{0.15\textwidth}
    \includegraphics[width=1.0\textwidth]{hardware_exps/placing-demo-2.png}
    \subcaption{\small{Demo}}
    \end{subfigure}
    \begin{subfigure}[b]{0.15\textwidth}
    \includegraphics[width=1.0\textwidth]{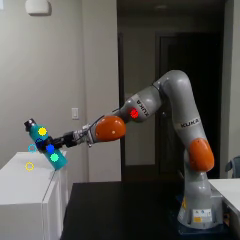}
    \subcaption{\small{Default}}
    \end{subfigure}
    \begin{subfigure}[b]{0.15\textwidth}
    \includegraphics[width=1.0\textwidth]{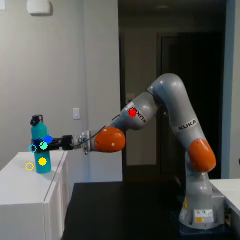}
    \subcaption{\small{Weighted}}
    \end{subfigure}
    \begin{subfigure}[b]{0.15\textwidth}
    \includegraphics[width=1.0\textwidth]{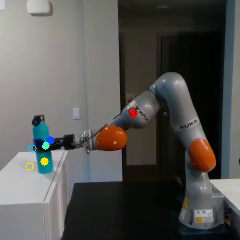}
    \subcaption{\small{TimeDep}}
    \end{subfigure}
    \begin{subfigure}[b]{0.15\textwidth}
    \includegraphics[width=1.0\textwidth]{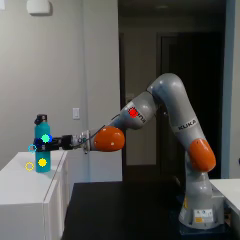}
    \subcaption{\small{RBF}}
    \end{subfigure}
\caption{
\small{Column a) Human Demonstration that is used for the IRL algorithm to extract cost functions. Column b)-d) Comparison of visual MPC result using the default and learned costs. First row corresponds to timestep $t=0$, middle row to $t=5$ and bottom row to $t=10$ of executing the placing task. The detected and goal keypoints in each image are depicted using filled and hollow circles respectively.}
}
\label{fig:placing_task_2}
\vspace{-0.5cm}
\end{figure}

Figure \ref{fig:placing_task_2} visually depicts the results for starting configuration 2.
